%% file: main-arxiv.tex
\documentclass[runningheads]{llncs}
\usepackage{eccv}
\usepackage{eccvabbrv}
\usepackage{graphicx}
\usepackage{booktabs}
\usepackage[accsupp]{axessibility}  
\usepackage{times}
\usepackage{epsfig}
\usepackage{graphicx}
\usepackage{amsmath}
\usepackage{amssymb}
\usepackage{nicefrac}
\usepackage{tabularx}
\usepackage{gensymb}
\usepackage{array}
\usepackage{tikz}
\usepackage{pgf-pie}
\usetikzlibrary{decorations, decorations.text,backgrounds}
\usepackage{pgfplots}
\usepackage{pgfplotstable}
\usepackage{booktabs}
\usepackage{array}
\usepackage{colortbl}
\usepgfplotslibrary{fillbetween}
\usepackage{pgfplotstable}
\pgfplotsset{compat=1.11,
        /pgfplots/ybar legend/.style={
        /pgfplots/legend image code/.code={%
        \draw[##1,/tikz/.cd,bar width=8pt,yshift=-0.2em,bar shift=0pt]
                plot coordinates {(0cm,0.8em)};},
},
}
\usetikzlibrary{patterns}
\usepackage[figuresleft]{rotating} 
\usepackage[pagebackref,breaklinks,colorlinks,citecolor=eccvblue]{hyperref}
\usepackage{orcidlink}
\usepackage{xcolor}
\usepackage{pifont}
\newcommand{\cmark}{\textcolor{Green}{\ding{51}}}%
\newcommand{\xmark}{\textcolor{red}{\ding{55}}}%
\usepackage{multirow} 
\usepackage{arydshln} 
\usepackage{siunitx} 
\newcolumntype{T}[1]{S[table-format=#1]}
\newcolumntype{U}[1]{S[table-format=#1,
                       round-mode=places, 
                       round-precision=2]}                       
\newcolumntype{R}{S[table-number-alignment=right,table-alignment-mode=none,table-text-alignment=right]}
\newrobustcmd\NoteMark[1]{%
    \textsuperscript{\emph{#1}}%
}
\usepackage{collcell} 
\begin{document}
\title{Bosch Street Dataset: A Multi-Modal Dataset with Imaging Radar for Automated Driving} 
\titlerunning{Bosch Street Dataset}
\author{Karim Armanious\inst{1} \and
Maurice Quach\inst{2} \and
Michael Ulrich\inst{2} \and
Timo Winterling\inst{1} \and
Johannes Friesen\inst{1} \and
Sascha Braun\inst{1} \and
Daniel Jenet\inst{1} \and
Yuri Feldman\inst{2} \and
Eitan Kosman\inst{2} \and
Philipp Rapp\inst{1} \and
Volker Fischer\inst{2} \and
Marc Sons\inst{2} \and
Lukas Kohns\inst{1} \and
Daniel Eckstein\inst{1} \and
Daniela Egbert\inst{1} \and
Simone Letsch\inst{1} \and
Corinna Voege\inst{1} \and
Felix Huttner\inst{1} \and
Alexander Bartler\inst{2} \and
Robert Maiwald\inst{1} \and
Yancong Lin\inst{3} \and
Ulf R{\"u}egg\inst{1} \and
Claudius Gl{\"a}ser\inst{2} \and
Bastian Bischoff\inst{2} \and
Jascha Freess\inst{1} \and
Karsten Haug\inst{1} \and
Kathrin Klee\inst{1} \and
Holger Caesar\inst{3}
}
\authorrunning{K.~Armanious, M. Quach, M. Ulrich et al.}
\institute{$^{1}$Robert Bosch GmbH, Cross-Domain Computing Solutions\footnote{Corresponding authors:
\email{karim.armanious@bosch.com},\email{kathrin.klee@de.bosch.com}, \email{maurice.quach@de.bosch.com}} \and \newline 
$^{2}$Bosch Corporate Research \and \newline 
$^{3}$Intelligent Vehicles Lab, Delft University of Technology
}
\maketitle
\begin{abstract}
  This paper introduces the Bosch street dataset (BSD), a novel multi-modal large-scale dataset aimed at promoting highly automated driving (HAD) and advanced driver-assistance systems (ADAS) research. Unlike existing datasets, BSD offers a unique integration of high-resolution imaging radar, lidar, and camera sensors, providing unprecedented 360-degree coverage to bridge the current gap in high-resolution radar data availability. Spanning urban, rural, and highway environments, BSD enables detailed exploration into radar-based object detection and sensor fusion techniques. The dataset is aimed at facilitating academic and research collaborations between Bosch and current and future partners. This aims to foster joint efforts in developing cutting-edge HAD and ADAS technologies. The paper describes the dataset's key attributes, including its scalability, radar resolution, and labeling methodology. Key offerings also include initial benchmarks for sensor modalities and a development kit tailored for extensive data analysis and performance evaluation, underscoring our commitment to contributing valuable resources to the HAD and ADAS research community.
\end{abstract}
\begin{figure}
\centerline{\includegraphics[width=0.99\columnwidth,trim={0 0 5cm 0},clip]{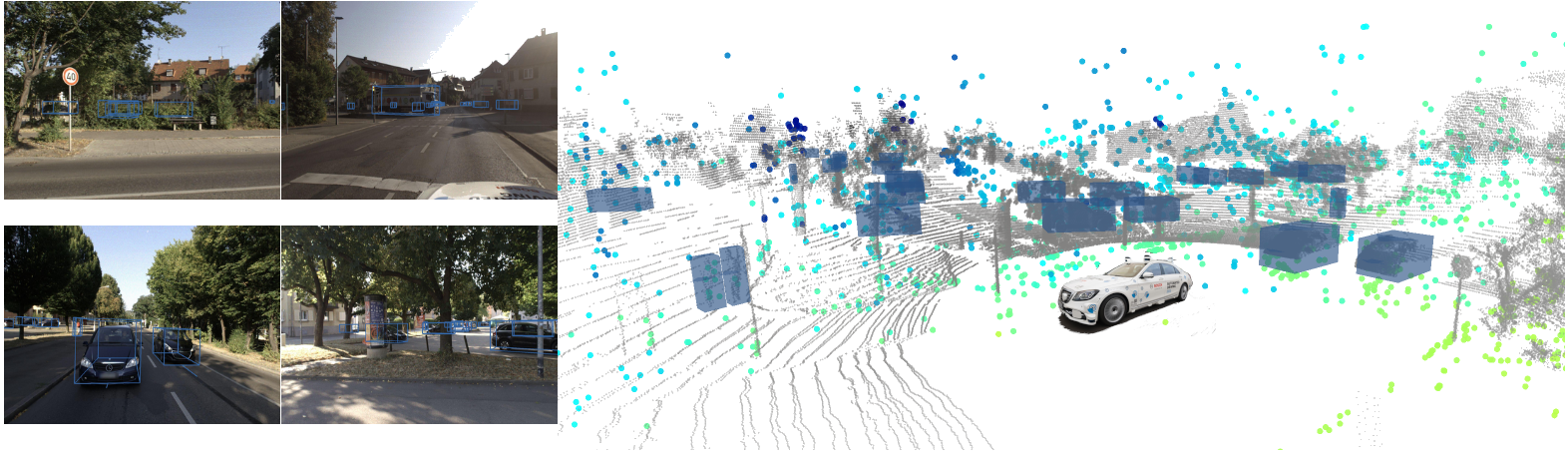}}
\caption{The Bosch street dataset (BSD) showcased with sample data captures. The setup includes 9 imaging radar sensors, 4 cameras, and a 64-channel lidar sensor, covering a complete 360-degree field of view. Data from varied settings —urban, rural, and highways — highlight the dataset's versatility across different geographical locations and under various weather and lighting conditions. The dataset features 13.6k sequences and over 36 hours of measurements, all annotated with bounding box labels to facilitate advanced HAD research.}
\label{fig:teaser}
\end{figure}
\section{Introduction}
Lidar, camera, and radar sensors are the most important modalities for highly automated driving (HAD) and advanced driver-assistance systems (ADAS). Every sensor comes with its distinct advantages, with radar sensors standing out for their robustness in diverse weather and illumination conditions, coupled with their cost-effectiveness and capability of efficiently estimating the velocity of the underlying target objects.
Substantial progress happened in radar sensing technology with imaging radar sensors, enabling the development of novel processing methodologies. 
Traditional radar sensors suffered from poor resolution, prohibiting the use of deep-learning based perception methods and resulting in limited performance improvements from sensor fusion. 
In contrast, modern imaging radar sensors capture more fine-grained information, such as the shape or extent of objects, allowing radar to be used in state-of-the-art sensor fusion methods, or even as a standalone modality. 

Several high-quality multi-modal datasets for autonomous driving have emerged, encompassing large-scale collections of data captured by powerful camera and lidar sensors. 
Regrettably, these datasets often lack high-resolution radar sensors, thereby limiting research possibilities involving radar.
For instance, the widely used ``nuScenes\,\cite{Caesar:2020}'' dataset uses only low-resolution radar sensors, while recent multi-modal datasets such as ``Argoverse~2\,\cite{Wilson:2023}'' and ``ZOD\,\cite{Alibeigi:2023}'' do not contain radar data at all. 
Furthermore, existing multi-modal datasets often focus on urban scenarios. 
While relevant, fully functional HAD must consider rural and highway scenarios as well, where radar excels at high velocities and far ranges. 
For these reasons, radar is already a key ingredient of today's driver assistance systems and with the steadily improving resolution of imaging radars will only increase in importance. 

The radar research community published several radar-focused datasets recently. 
However, these datasets often suffer from limitations in terms of scale, scene diversity, and multi-modal coverage, featuring only limited camera and lidar data. 
For example, datasets like ``View~of~Delft\,\cite{Palffy:2022}'' and ``TJ4DRadSet\,\cite{Zheng:2022}'', contain only a couple of dozen sequences and featuring only a single radar, lidar and camera sensors.

This work introduces a novel ADAS-focused dataset tackling the prior shortcomings of its predecessors. Namely, the introduced dataset is a fully multi-modal dataset, collected using a sensor-belt of state-of-the-art imaging radar sensors, high resolution lidar scanners, and camera sensors providing a full 360-degree field-of-view (FoV). The dataset is the largest of its kind with far-range bounding box labels and high scene diversity including urban, rural, and highway conditions recorded from diverse geographical locations. The contributions of this work are outlined as follows:


\begin{itemize}
\let\labelitemi\labelitemii
    \item We introduce the Bosch street dataset (BSD), a multi-modal, large-scale, far-range, and diverse dataset including imaging radar. We hope BSD will serve as a cornerstone for establishing academic and research collaborations between Bosch and interested potential partners. 
    \item The dataset was collected using a state-of-the-art sensor belt of 9 imaging radars, 4 lidars and 4 cameras, offering full 360-degree coverage. 
    \item 3D bounding box labels are provided, making this the largest labeled object detection dataset of its kind. 
    \item We provide initial object detection baselines for each sensor modality and demonstrate that the previously observed performance gaps between radar and the other two modalities are significantly reduced in our dataset.
    \item We provide a comprehensive development kit that enables users to read, aggregate, visualize, perform object detection baselines, and evaluate on the dataset.
    \item We present statistics and ablation studies for the dataset.
\end{itemize}

\noindent More precisely, the findings of this paper are the following: 
\begin{itemize}
\let\labelitemi\labelitemii
    \item Training with $100$x more data in comparison to previous imaging radar datasets improves object detection performance by a significant factor (i.e., by a factor of at least 6). 
    \item Using an $8-20$x higher radar resolution in comparison to previous large-scale multi-modal datasets improves the performance significantly (i.e., by a factor of at least 3).
    \item We employ a combination of automated and manual labeling to ensure the accuracy and reliability of the labels. Experiments show better object detection performance when including the combination of both label types in comparison to training solely on a smaller set of manual labels. This indicates the solid labeling quality of our included auto-labels and highlights auto-labeling as a key enabling factor for scaling HAD-focused datasets.
\end{itemize}

In conclusion, this work addresses the pressing need for a large-scale dataset which includes high-resolution imaging radar. 
By introducing BSD, we provide researchers with a powerful tool to advance the fields of HAD and ADAS. 
Focusing on object detection for the moment, the dataset could be extended to further autonomous driving tasks in the future. 
We firmly believe that this dataset will pave the way for new breakthroughs in perception and safety, ultimately propelling us closer to a future where autonomous driving is the norm.

\begin{sidewaystable}
\center
\setlength\tabcolsep{1.43pt} 
\begin{tabular}{@{}l T{4.0\NoteMark{*}} R@{\,}l T{2.2} T{1} R@{\,}l r R@{\,}l r R@{\,}l R@{\,}l R@{\,}l rr T{>3{**}} T{>3} T{1} T{1} @{}}
\rotatebox{75}{\textbf{Dataset}} & 
\rotatebox{75}{\textbf{Year}} &
\rotatebox{75}{\textbf{Scenes}} & &
\rotatebox{75}{\textbf{Size (hr)}} & 
\rotatebox{75}{\textbf{\# Camera}} & 
\rotatebox{75}{\textbf{RGB imgs}} & &
\rotatebox{75}{\textbf{\# LiDAR}} & 
\rotatebox{75}{\textbf{Lidar PCs}} & &
\rotatebox{75}{\textbf{\# Radar}} &
\rotatebox{75}{\textbf{Radar PCs}} & & 
\rotatebox{75}{\textbf{Ann. frames}} & &
\rotatebox{75}{\textbf{3D boxes}} & &
\rotatebox{75}{\textbf{Night / Rain}} & 
\rotatebox{75}{\textbf{Urban / Highway}} & 
\rotatebox{75}{\textbf{FoV (deg)}} & 
\rotatebox{75}{\textbf{Range (m)}} & 
\rotatebox{75}{\textbf{Continents}} & 
\rotatebox{75}{\textbf{Cities}}\\
\midrule
KITTI \cite{Geiger:2013}              & 2012             & 22    &                         & 1.5   & 4x & 15   & k & 1x 64ch        & 15   & k & {---} & {---} &   & 15    & k & 200    & k & \xmark/\xmark & \cmark/\cmark & 180                       & <100 & 1 & 1\\  \hdashline
Waymo Open$^{\dag}$ \cite{Sun:2020}   & 2019\NoteMark{*} & 1.2   & k                       & 6.4   & 5x & 1.2  & M & 1x (+4x)*      & 230  & k & {---} & {---} &   & 230   & k & 12     & M & \cmark/\cmark & \cmark/\xmark & 360                       & <100 & 1 & 3\\  \hdashline
Once \cite{Mao:2021}                  & 2021             & 20    & \NoteMark{$\dag \dag$}  & 2.3   & 7x & 112  & k & 1x 40ch        & 16   & k & {---} & {---} &   & 16    & k & 417    & k & \cmark/\cmark & \cmark/\xmark & 360                       & <100 & 1 & {---}\\ \hdashline
Argoverse 2 \cite{Wilson:2023}        & 2023             & 1     & k\NoteMark{$\dag \dag$} & 4.2   & 9x & 2.7  & M & 2x 32ch        & 300  & k & {---} & {---} &   & 150   & k & 11.2   & M & \cmark/\cmark & \cmark/\xmark & 360                       & <100 & 1 & 6\\  \hdashline
ZOD$^{\ddagger}$ \cite{Alibeigi:2023} & 2023             & 1.5   & k                       & 8.2   & 1x & 295  & k & 1/2 x 128/16ch & 88   & k & {---} & {---} &   & 295   & k & 1.4    & M & \cmark/\cmark & \cmark/\cmark & 120$^{\ddagger}$          & <200 & 1 & 6\\  \midrule
nuScenes \cite{Caesar:2020}           & 2019             & 1     & k                       & 5.5   & 6x & 1.4  & M & 1x 32ch        & 400  & k & 5x 3D & 1.3   & M & 200   & k & 7.0    & M & \cmark/\cmark & \cmark/\xmark & 360                       & <100 & \textbf{2} & 2\\  \hdashline
Astyx \cite{Meyer:2019}               & 2019             & {---} &                         & 0.02* & 1x & 0.5  & k & 1x 16ch        & 0.5  & k & 1x 4D & 0.5   & k & 0.5   & k & 3.1    & k & \xmark/\xmark & \xmark/\cmark & 180                       & <100 & 1 & 1\\  \hdashline
DENSE \cite{Bijelic:2020}             & 2019             & {---} &                         & 0.4*  & 4x & 54   & k & 2x 64ch        & 27   & k & 1x 3D & 13.5  & k & 13.5  & k & 100    & k & \cmark/\cmark & \xmark/\cmark & 180                       & <100 & 1 & \textbf{14}\\ \hdashline
Zendar \cite{Mostajabi:2020}          & 2020             & 27    &                         & 0.14  & 1x & 11.4 & k & 1x 16ch        & 4.8  & k & 1x 3D & 4.8   & k & 4.8   & k & 410    & k & \xmark/\xmark & \cmark/\xmark & 180                       & <100 & 1 & 1\\  \hdashline
aiMotive** \cite{Matuszka:2022}       & 2022             & 176   &                         & 0.74  & 4x & 425  & k & 1x 64ch        & 26.5 & k & 2x 4D & 95.4  & k & 26.5  & k & 427    & k & \cmark/\cmark & \cmark/\cmark & 360                       & <200 & 2 & 3\\  \hdashline
View of Delft \cite{Palffy:2022}      & 2022             & 22    &                         & 0.25  & 1x & 26.1 & k & 1x 64ch        & 8.7  & k & 1x 4D & 11.3  & k & 8.7   & k & 123    & k & \xmark/\xmark & \cmark/\xmark & 180                       & <50 & 1 & 1\\  \hdashline
TJ4DRadSet \cite{Zheng:2022}          & 2022             & 44    &                         & 0.22  & 1x & 23.2 & k & 1x 32ch        & 7.7  & k & 1x 4D & 11.6  & k & 7.7   & k & 33.5   & k & \cmark/\xmark & \cmark/\xmark & 120$^{\ddagger \ddagger}$ & <100 & 1 & 1\\  \hdashline
Dual Radar \cite{Zhang:2023}          & 2023             & 151   &                         & 0.8   & 1x & 10   & k & 1x 80ch        & 10   & k & 2x 4D & 40    & k & 10    & k & 103.2  & k & \cmark/\cmark & \cmark/\xmark & 120                       & <100 & 1 & 1\\  \hdashline
\midrule
\textbf{Ours}                         & \textbf{2024}    & \textbf{13.6}  & \textbf{k}     & \textbf{36.5} & 4x & \textbf{5.2} & \textbf{M} & \textbf{4x 64ch} & \textbf{5.2}   & \textbf{M}  & \textbf{9x 4D} & \textbf{17.7}   & \textbf{M}  & \textbf{1.3}   & \textbf{M} & \textbf{32.1}   & \textbf{M} & \textbf{\cmark/\cmark} & \textbf{\cmark/\cmark} & \textbf{360}                       & \textbf{<200} & \textbf{2} & 9 \\ \bottomrule
\end{tabular}
\caption{
Annotated HAD datasets comparison. ($^{\dag}$) We report numbers only for the annotated Waymo Open Perception dataset, 2030 scenes are available in total but only 1200 are annotated.  * Channel count not specified: 1x mid-range lidar, 4x short-range lidar. ($^{\dag \dag}$) We report only number of annotated scenes. ** Similarly a mixture of auto and manual labels. ($^{\ddagger}$) We report numbers for the ZOD sequence datasets with 20s scenes, the 120\degree FOV represents the camera coverage of this dataset. ($^{\ddagger \ddagger}$) FoV of the radar sensor.
}
\label{tab:dataset_comparison}
\end{sidewaystable}

\section{Related Work}

State-of-the-art HAD datasets such as ``Waymo Open\,\cite{Sun:2020}'', ``Once\,\cite{Mao:2021}'', ``Argoverse~2\,\cite{Wilson:2023}'', and ``Zenseact Open Dataset\,\cite{Alibeigi:2023}'' do not feature radar at all. While several existing automotive datasets either contain data from high-resolution radar sensors \cite{Meyer:2019, Matuszka:2022, Palffy:2022, Zheng:2022} or contain a sufficient number of annotated frames \cite{Sun:2020, Wilson:2023, Caesar:2020},
top-tier research needs both: sensors with sufficient resolution and sufficient dataset size. 

Furthermore, many existing datasets are highly specialized, focusing on one single aspect.
This hinders comparability of algorithms. 
For instance, some datasets lack high resolution lidar sensors \cite{Mao:2021, Wilson:2023}, others only have a single lidar \cite{Geiger:2013, Sun:2020, Mao:2021, Caesar:2020, Meyer:2019, Mostajabi:2020, Matuszka:2022, Palffy:2022, Zheng:2022, Zhang:2023}, and some only rely on a single camera \cite{Alibeigi:2023, Meyer:2019, Mostajabi:2020, Palffy:2022, Zheng:2022, Zhang:2023}.

A pioneering multi-modal dataset is nuScenes~\cite{Caesar:2020}, with radar sensors, large scale ($40$k annotated frames), and diverse sequences. 
However, compared to BSD's high-resolution imaging radars, nuScenes's radars are low resolution and provide no elevation data.
Specifically, detections of cars in BSD contain $22.47$ radar points on average while the nuScenes average is $1.89$ points ($11.88\times$ lower).
Consequently, few radar-only submissions were made to the nuScenes leaderboard \cite{Svenningsson:2021,Ulrich:2022,Lippke2023}, 
despite nuScenes's popularity for lidar and camera perception. 
Even then, radar-only submissions typically exhibit poor detection performance compared to camera and LiDAR.
Medium-scale datasets for multi-modal perception, such as ``PixSet\,\cite{Deziel:2021}'' and ``aiMotive\,\cite{Matuszka:2022}'', also contain radar data.
However, these radar sensors also have a low resolution, making them unsuitable for radar perception research. 

``Astyx \cite{Meyer:2019}'' is a high-resolution radar dataset, which is very small with only 546 frames. 
The remaining automotive radar perception datasets neglected broad applicability and scale, choosing instead to focus on specific aspects such as multi-static radar in Pointillism\,~\cite{Bansal:2020}, synthetic aperture radar in ``Zendar\,\cite{Mostajabi:2020}'', a comparison of different radar types in ``Dual Radar\,\cite{Zhang:2023}'', and adverse weather conditions in the ``DENSE'' (project) dataset\,\cite{Bijelic:2020}. ``View~of~Delft\,\cite{Palffy:2022}'' and ``TJ4DRadSet\,\cite{Zheng:2022}'' use modern, high-resolution radar sensors.  They provide significantly more annotated frames than previous datasets with high resolution sensors. 
Yet, the number of annotated frames is still smaller than popular automotive datasets such as nuScenes and Waymo Open (Tab.~\ref{tab:dataset_comparison}). 

Furthermore, perception on radar spectra is an active research area.
The radar spectrum is a low-level representation of the measurements before the point cloud (PC) is derived. 
Some of the most relevant radar spectra datasets are ``K-Radar~\cite{Paek:2022}'', and ``Radial~\cite{Rebut:2022}''. Besides object detection and tracking, radar datasets exist for localization \cite{Barnes:2020}, semantic segmentation of radar point clouds~\cite{Schumann:2021}, ghost object classification \cite{Kraus:2021}, and perception in bad weather~\cite{Sheeny:2021}. 
A comprehensive overview is given in~\cite{Zhou:2022}. 

BSD is the first radar-centric dataset providing both substantial number of annotated frames as well as high-resolution radar sensors (Tab.\,\ref{tab:dataset_comparison}). 
We follow in the footsteps of nuScenes~\cite{Caesar:2020}, but with modern radar sensors and even larger scale. 
Further, we are convinced that our powerful setup with four cameras, four $64$-layer lidars,
and nine radars is a key ingredient for research in multi-modal perception. 

\section{Background on Radar Sensing}
Radar is a unique modality, with new possibilities and challenges for the research community. 
In this section, we elaborate the differences to lidar and camera to underline the need for research in this domain. 

\subsection{Radial Velocity}
The availability of radial relative velocity (Doppler) measurements for each reflection point is the most prominent radar feature.
Strictly speaking, it is not a property of the radar modality, but rather a result of the frequency modulation scheme that is usually
employed in automotive radar sensors.
Doppler is already a powerful feature in classical radar perception, which highlights its usefulness. 
For example, moving objects are often of high interest.
They can be identified directly by their Doppler measurement, if their motion contains a sufficient radial component. 
The radial velocity greatly facilitates estimating the dynamic state of objects.

\subsection{Fluctuating Point Cloud}
Typical automotive radar sensors operate at a carrier frequency~$f_\mathrm{c}$ of~$76\,\text{GHz}$ to $78\,\text{GHz}$, which corresponds to a wavelength of $\lambda \approx 4\,\text{mm}$. 
The roughness of most surfaces, such as roads or objects of interest, is smoother than this, resulting in mirror-like reflections. 
This results in highly fluctuating intensities for small changes in aspect angle~\cite{Richards:2014,Ludloff:2002}.
These fluctuations result in different point clouds of the same object, in slightly different frames. 
This makes object detection in radar point clouds challenging.

\subsection{Separability of Radar Points}
The wavelength determines the size of the transmitter and receiver elements. 
Larger antenna elements (aperture), relative to wavelength, allow for a narrower beam and higher spatial selectivity. 
Contrary to radar sensors, lidar sensors scan the scene with narrow beams, which are possible due to the small wavelength of near-infrared light. 
In contrast, a limited radar sensor size of a few centimeters can focus the radar beam only to a tenth of a degrees. 
Hence, radar illuminates the whole field of view at once and reflection points are \emph{separated} only later in the digital signal processing.

As a result, radar possesses resolution mainly in the range (distance) and Doppler dimension, which means that separate points are generated on an object, when it is either extended in range or moving with different radial velocities. 
Subtle differences in radial velocity in the order of $0.1\,\nicefrac{\text{m}}{\text{s}}$, for instance, are enough to separate reflections on different object parts. 
Specifically, radial velocity can differ for different parts of a rigid object, due to different observation angles for different parts. 
In contrast, lidar and camera resolve points mainly through azimuth and elevation angle dimensions and have range and Doppler only as features. 
This has the following implications:
\begin{itemize}
    \item Few radar points for problematic objects: Some objects might generate only a few radar points, when they have no significant extent in range or Doppler dimension. To give an example, this is often observed with stationary pedestrians. 
    \item Angular ambiguities: Radar sensors are often subject to angular ambiguities which have to be resolved.
    \item Even more ambiguities: Multiple radar points may be positioned at the exact same location with different radial velocities. For instance, a bicycle can typically result in several points as its different parts have different radial velocities.
    \item Suitable for far distances: The whole scene is illuminated at once with the radar beam. 
    This means that objects are missed seldomly, which makes radar suitable for high distances.
    In contrast, an object may lie between neighboring lidar beams and be overlooked. 
    In addition, all reflection points of one measurement correspond to the same time. 
    There are no shutter effects as with camera or distortion due to the rotating scans of spinning LiDARs.  
\end{itemize}

\subsection{Radar for Redundancy or as a Cost-Efficient Alternative to Lidar}
As an active sensor, radar is independent of external lighting conditions, similar to lidar. 
Furthermore, radar is less sensitive to adverse weather conditions, such as fog or rain, in comparison to lidar. 
This makes radar an excellent choice to add redundancy when lidar and camera sensors fail.

Alternatively, radar could be a cost-efficient substitute for lidar:
 the cost of one of the lidar sensors of this dataset is more than two orders of magnitude higher than the commercial variant of the included  radar sensor. 
Radar-only or camera-radar setups could democratize HAD, bringing state-of-the-art research to the broader public, while the industry is still struggling to make lidar-based systems commercially successful. 

%
%
\section{The Bosch Street Dataset}

\subsection{Vehicle and Sensor Setup}
\begin{figure}[!tb]
    \centering
    \begin{subfigure}{0.49\columnwidth}
        \includegraphics[width=1.0\columnwidth]{%
            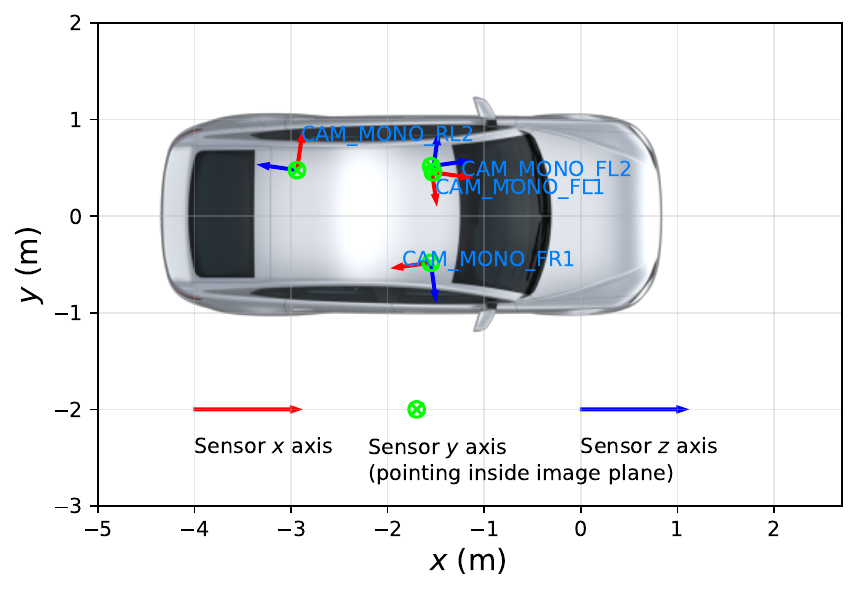}
        \caption{}
        \label{fig:vehicle:sensor_setup:cameras}
    \end{subfigure}
    \begin{subfigure}{0.49\columnwidth}
        \includegraphics[width=1.0\columnwidth]{%
            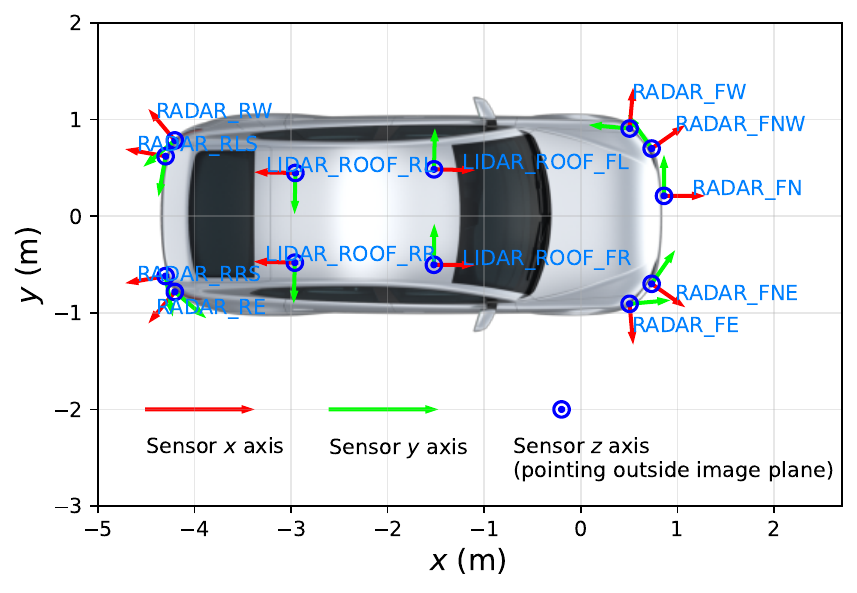}
        \caption{}
    \end{subfigure}
    \caption{Bird's eye view of the sensor setup.
            (a) shows the cameras and (b) shows the lidars and the radars.
            The sensors are depicted by means of of their local coordinate frames.
            Axes according to ISO~8855.
            Vehicle not to scale.
    }
    \label{fig:vehicle:sensor_setup}
\end{figure}

We use a fleet of five vehicles for capturing the data.
Four of those vehicles are located in Germany, and one vehicle is located in the USA (California).
Every vehicle has a sensor setup consisting of nine imaging radar prototypes,
four lidar sensors, and four mono cameras.
Additionally, an inertial measurement unit is used in conjunction with wheel speed sensors to provide odometry information by means of dead reckoning.
This allows us to define an inertial coordinate system that does not suffer from GPS outage
or discontinuous corrections. Fig.~\ref{fig:vehicle:sensor_setup}
shows the sensor mounting poses of the sensors.

\subsection{Sensor Specifications}

\subsubsection{Radar Sensor Specifications and Features.}
The capturing vehicle is equipped with a 360\degree~radar belt consisting of
nine imaging radar prototypes.
Those are long-range radars that employ chirp-sequence frequency modulation and provide up to $N_\text{pts}=1024$ reflection points per scan per sensor
at a sampling interval of~$\Delta t_\text{rad} = 66\,\text{ms}$ ($15\,\text{Hz}$).
The radar prototype is derived from a high-trim automotive radar sensor.
This sensor provides an unambiguous Doppler measurement, a high spatial resolution, and a high angular resolution.


Every reflection point has the following attributes:
radial distance~$r$, radial relative velocity~$\dot{r}=v_{\text{rad},\text{rel}}$ (Doppler),
azimuth angle~$\theta$, elevation angle~$\varphi$,
and radar cross section~$\sigma$.
A post-processing step compensates the contribution of the ego-vehicle movement $\mathbf{v}_\text{ego}\in\mathbb{R}^3$ to the measured radial velocity in order to obtain the radial velocity over ground
\begin{equation}
    v_{\text{rad},\text{og}} = v_{\text{rad},\text{rel}} + \mathbf{d}^T \cdot \mathbf{v}_\text{ego},
\end{equation}
where $\mathbf{d}\in\mathbb{R}^3$ is a normalized vector ($||\mathbf{d}||_2=1$) pointing from the sensor origin to the reflection point.
Furthermore, the distance and angles are converted into Cartesian coordinates.

\subsubsection{Lidar Specifications.}
We utilize four modified 64-channel lidar sensors operating
at a wavelength of 905 nm.
These sensors offer a vertical field of view
of 16 degrees (-8 to 8 degrees) with a constant resolution of 0.25 degrees across 64 channels.
Additionally, we limit the horizontal field of view to 150 degrees.
We provide the Cartesian coordinates
of the measured lidar point cloud (given in sensor coordinates),
the intensity~$I$ (ranging from 0.0 to 1.0), as well as a per point timestamp that is relative to the start of the lidar scan.
The sample rate of all lidar sensor measurements is~$10\,\text{Hz}$.

\subsubsection{Camera Specifications.}
We provide the video signals of four roof-mounted mono cameras.
The images come at a frequency of~$10\,\text{Hz}$ and with a resolution of~$1920 \times 1090$ pixels.
The camera heads being used for the BSD were selected such that the optical axes of the front and back facing cameras are approximately aligned,
similar for the left and right facing cameras, see Fig.~\ref{fig:vehicle:sensor_setup:cameras}.


\subsubsection{Odometry Specifications.}
Odometry data is provided as the vehicles' positions~$x$, $y$, and $z$ with respect to an inertial coordinate system,
velocity~${v_{\text{x}}}$, ${v_{\text{y}}}$, ${v_{\text{z}}}$ with respect to an inertial coordinate system and given
in coordinates of the ego vehicle coordinate system,
as well as roll~$\theta$, pitch~$\phi$ (with respect to the road), yaw~$\psi$ (with respect to the inertial coordinate system) and the angular velocity (yawrate) over ground~$\omega$.
The frequency is~$100\,\text{Hz}$.
 
\subsection{Data Description}
The data is provided in the form of $13\,649$ sequences (9431 train, 2828 val, 1390 test), to allow for temporal processing. 
Each sequence is between 5 and 10 seconds long.
The aggregated duration of the dataset is 36.5 hours (25.2h training, 7.6h validation, 3.7h testing) containing 33.9h of automatically and 2.6h of manually labeled data. 
 
The sensor data as described in the sections above are provided ``as is'' and kept as raw as possible on purpose,
except for their conversion into Cartesian coordinates.
This is to allow for most flexibility when using the BSD. 
Common preprocessing steps are provided in the development kit accompanying the data. 
In addition to the sensor data, we provide labels in the form of 3D oriented bounding boxes,
each described by its center position~$\mathbf{r}=\left( x, y, z \right)$, extent $(L, W, H)$, orientation angle $\psi$, and velocity~$\mathbf{v}=\left( v_x, v_y \right)$. 
More details about the actual labeling process can be found in Sec.~\ref{sec:labeling}.

In addition to the sensor data and labels,
we also provide \emph{sequence metadata}.
This metadata includes the sensor mounting poses,
represented as homogeneous matrices~$T_i \in \mathbb{R}^{4 \times 4}$,
and intrinsic calibration matrices for all cameras.

\subsection{Labeling Process}
\label{sec:labeling}
The scale of the dataset plays a crucial role, as will be verified empirically in Sec.\,\ref{sec:experiment_dataset_size}.
Therefore, we include labels originating from our auto-labeling approach alongside labels originating from manual labeling.
The former enables efficient label generation, whereas the latter, albeit more time-consuming and costly, yields labels of superior quality with regards to the scale and orientation of the bounding boxes.

In the BSD, 94,212 frames are manually annotated (63,134 training, 9,296 validation, 21,782 test).
Those are complemented by an additional number of 1,220,162 frames that are auto-labeled (845,308 training, 263,902 validation, 110,952 test).

\subsubsection{Manual Labeling.}
The manual labels are created by experienced labeling experts.
Labeling is done in a ``multi sensor labeling'' approach,
that is, the 3D oriented bounding boxes are created leveraging the dense lidar point clouds at a frequency of~$10\,\text{Hz}$
and then verified by looking at the radar and camera measurements.

\subsubsection{Auto-Labeling.}
The auto-labeling system utilizes a high-performance multi-modal tracking and fusion system,
specifically the Bernoulli Gaussian Sum Filter~\cite{ristic:2013}.
As we are processing the data offline, we can implement this filter in an acausal manner
which enables us to do additional processing steps that cannot be done in online
processing.
This includes doing a backward-forward pass (in time),
leveraging track termination from the backward pass to enhance track initiation in the forward pass.
As a result, it minimizes track losses, reduces noise in state estimates,
and maintains a consistent (i.e., constant) spatial extent.
The labels generated through this auto-labeling process undergo additional post-processing 
and quality checks (such as for instance removing implausible tracks).
Radar is used as reference sensor for auto-labeling, thus resulting in a label frequency of~$15\,\text{Hz}$.

Auto-labeled sequences are manually curated to ensure high label quality and reliability.
Specifically, in case manual curation of a sequence uncovers major issues, the sequence length is adjusted while ensuring that it remains longer than 5 seconds.
That is, we discard low quality sequences and sequences shorter than 5 seconds. 

\begin{figure}
    \centering
    \includegraphics[width=.32\textwidth]{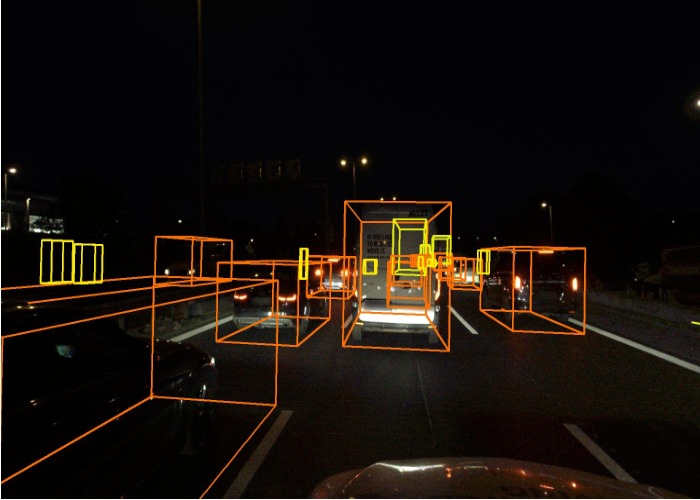}
    \includegraphics[width=.32\textwidth]{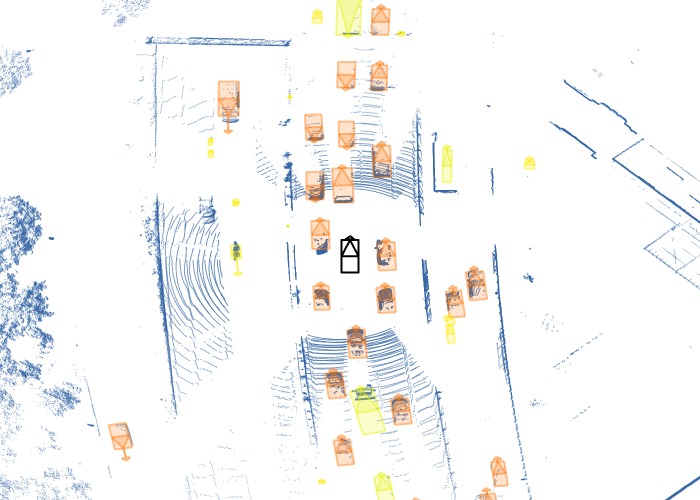}
    \includegraphics[width=.32\textwidth]{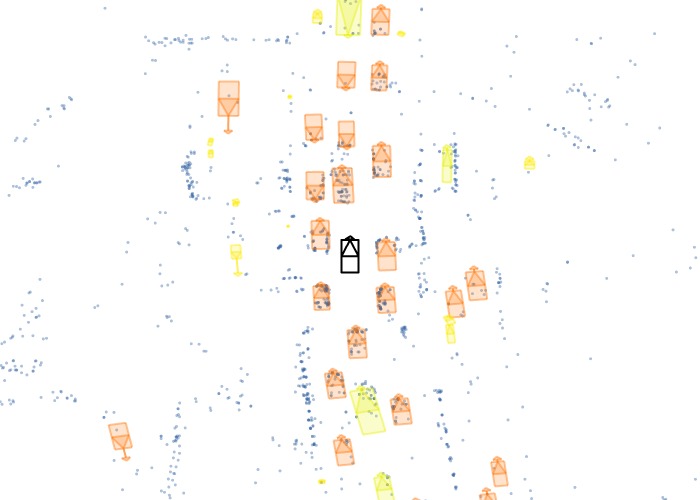}

    \includegraphics[width=.32\textwidth]{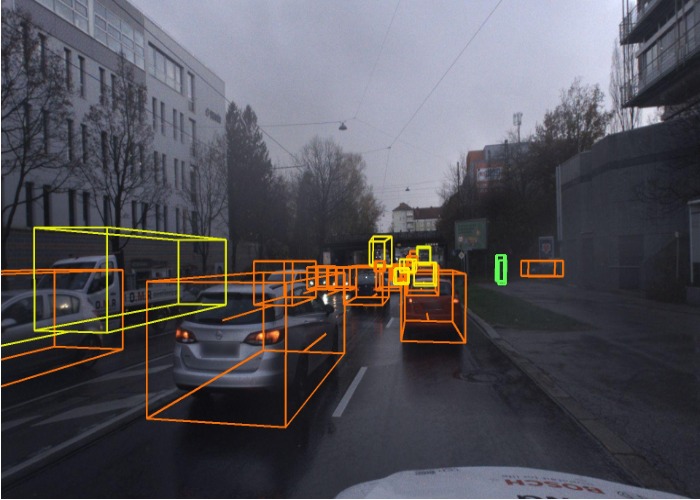}
    \includegraphics[width=.32\textwidth]{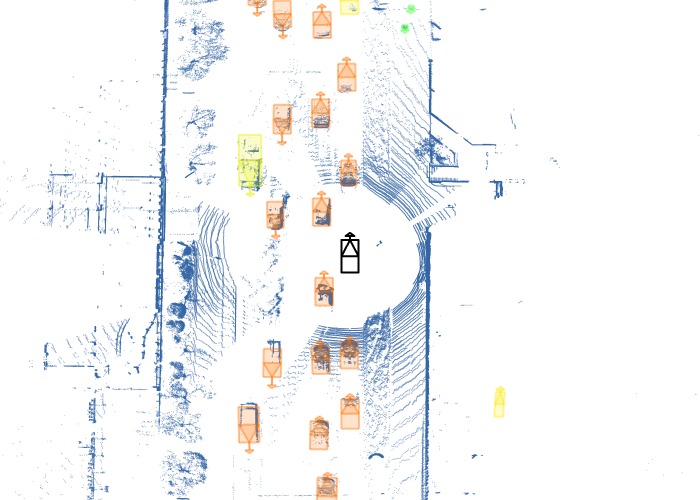}
    \includegraphics[width=.32\textwidth]{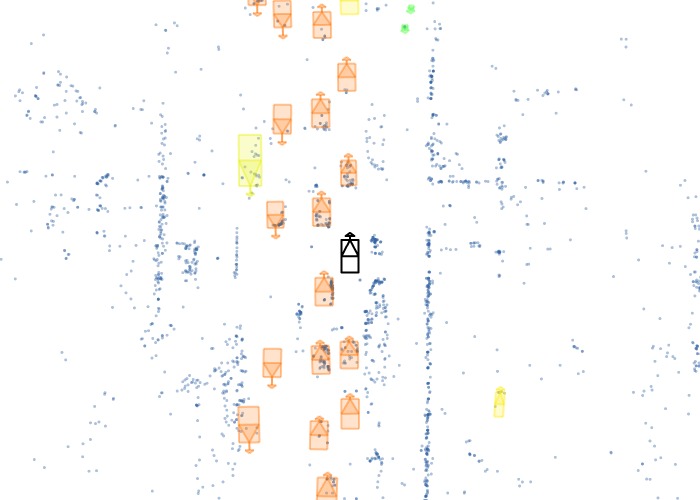}

    \includegraphics[width=.32\textwidth]{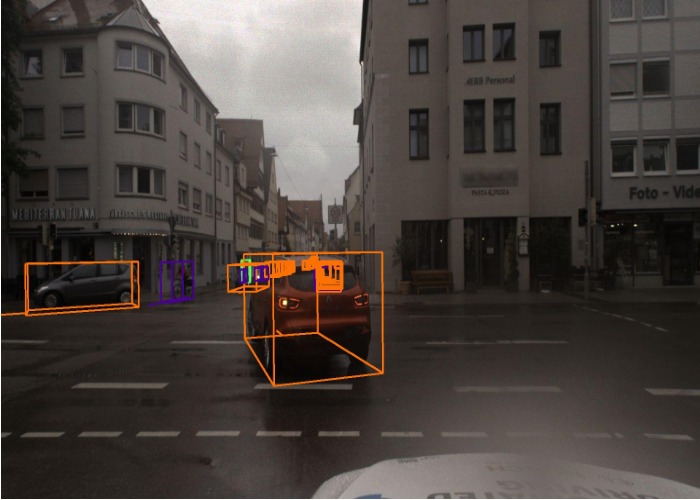}
    \includegraphics[width=.32\textwidth]{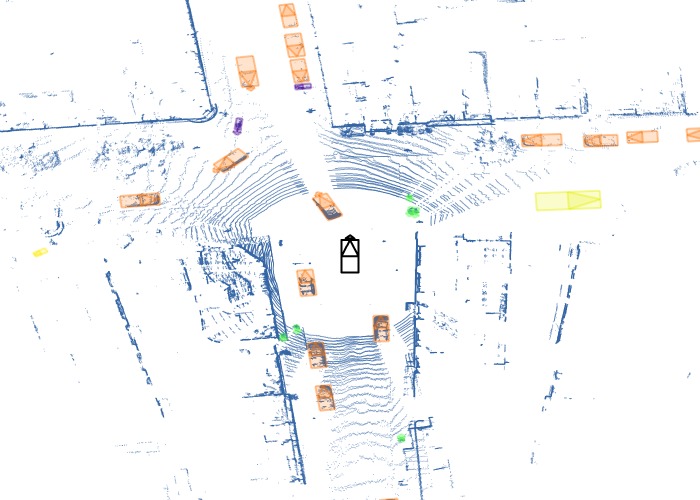}
    \includegraphics[width=.32\textwidth]{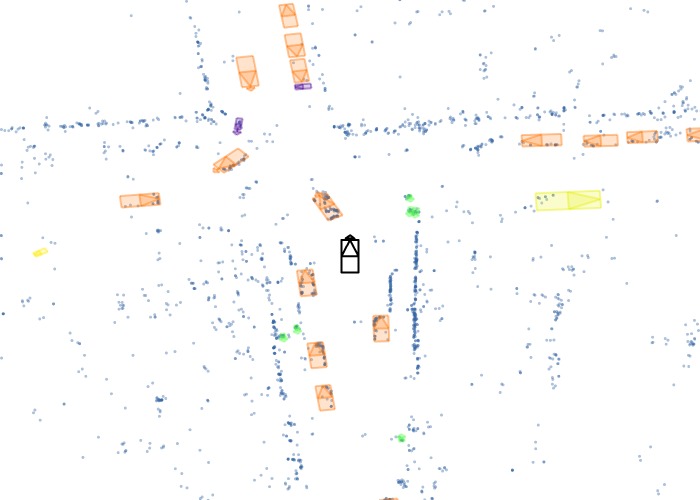}

    \includegraphics[width=.32\textwidth]{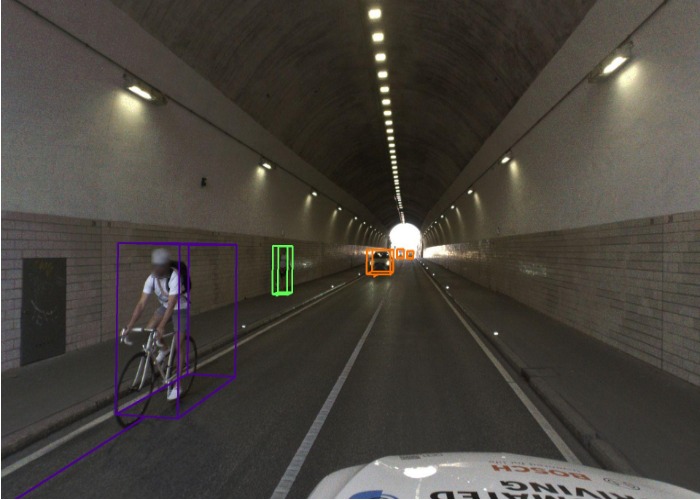}
    \includegraphics[width=.32\textwidth]{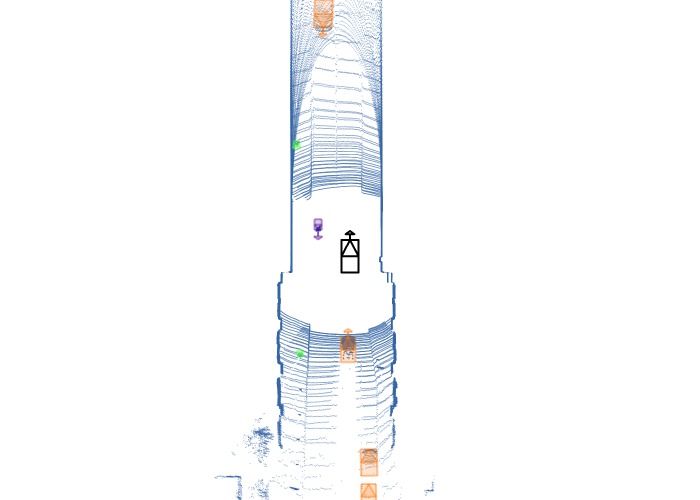}
    \includegraphics[width=.32\textwidth]{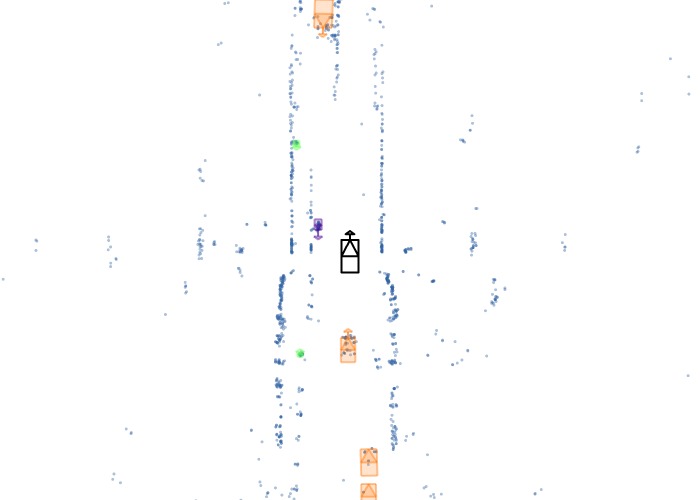}

    \includegraphics[width=.32\textwidth]{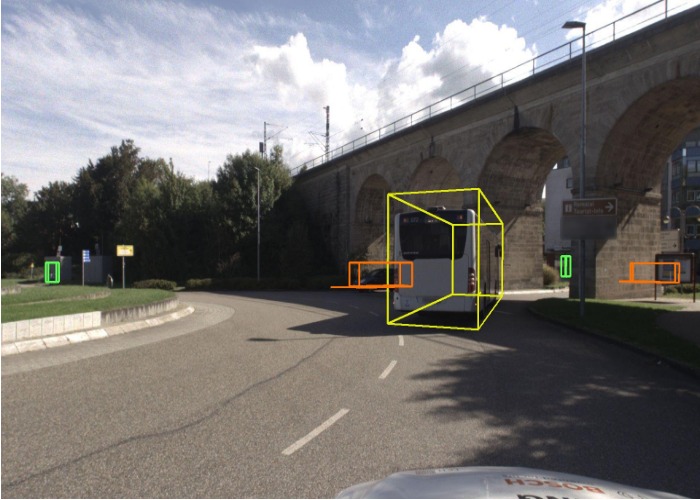}
    \includegraphics[width=.32\textwidth]{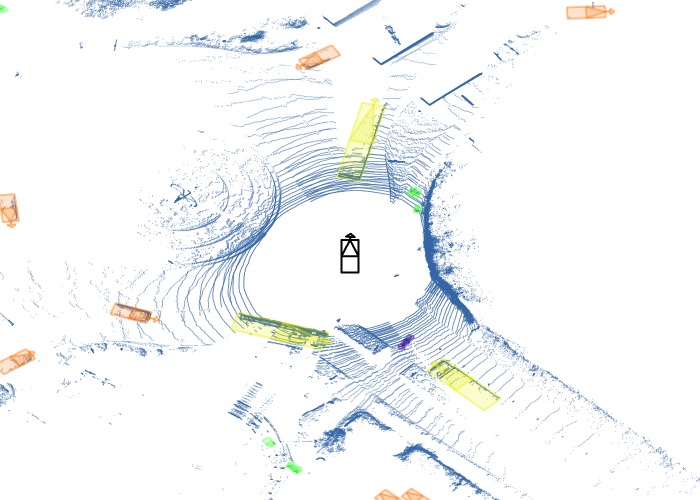}
    \includegraphics[width=.32\textwidth]{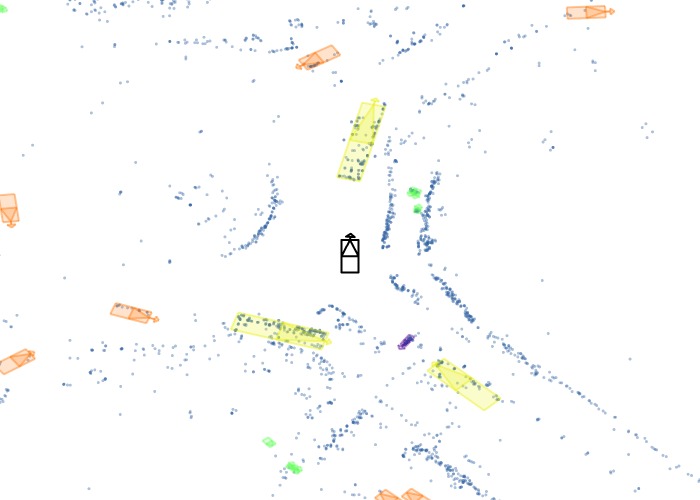}
    \caption{Example scenes from the dataset. The first column shows the camera image, the middle column lidar, and the right column radar. The rows represent various environmental conditions from top to bottom: a nighttime scene, a rainy scene, an intersection, a tunnel, and a roundabout.}
    \label{fig:dataset_examples}
\end{figure}

\subsubsection{General Remark.}
Since auto-labeling and manual labeling lead to different label frequencies,
we interpolate labels onto a common label frequency of~$20\,\text{Hz}$.

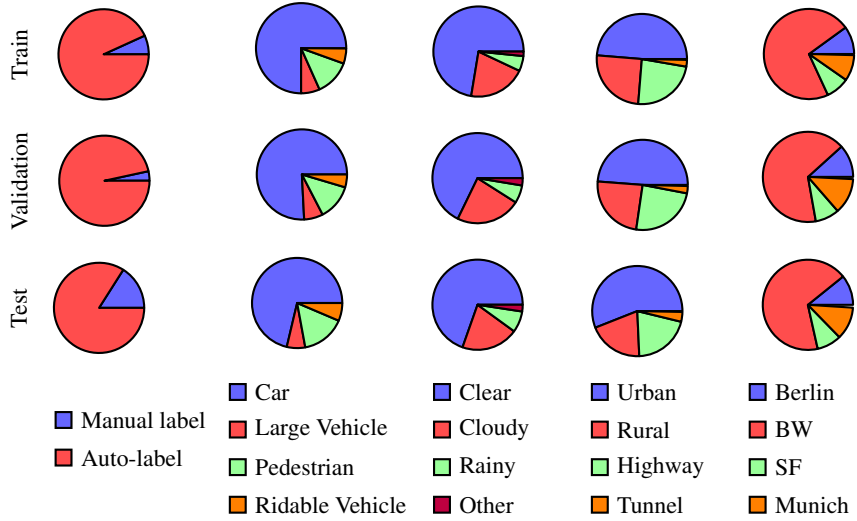
\begin{figure}
    \input{figures/dataset_statistics_piecharts.tikz}
    \caption{Distribution of the train, validation, and test split of the dataset. The dataset is diverse with respect to location, weather, time of day, and driving situation. BW stands for Baden-W\"urttemberg and SF stands for San Francisco Bay area. The test dataset contains more manually annotated data, to ensure highest data quality. 
    }
    \label{fig:data_piechart}
\end{figure}
\begin{figure}
    \begin{subfigure}[b]{0.485\textwidth}
    \input{figures/labels_over_range.tikz}
    \end{subfigure}
    \quad
    \begin{subfigure}[b]{0.485\textwidth}
    \input{figures/labels_over_velo.tikz}
    \end{subfigure}
    \caption{Distribution of labels (for the classes also shown in Fig.~\ref{fig:data_piechart}) over distance (range) on the left and velocity on the right comparing the BSD dataset to nuScenes. Density on the y-axis was computed as histograms, normalized to have an area of 1. This allows to compare coverage and shows that our BSD also provides instances for distant labels, i.e. beyond 100 m, and fast labels, i.e. beyond 20 m/s.
    }
    \label{fig:data_range_vel_dist}
\end{figure}
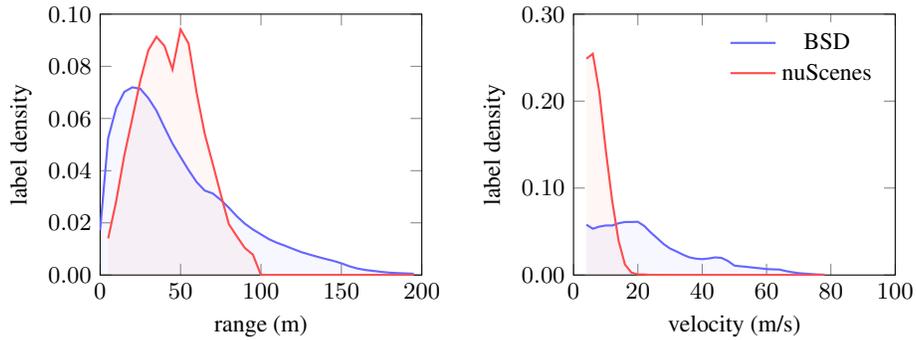
\begin{figure}
    \centering
    \input{figures/statistic_point_cloud_density_test.tikz}
    \caption{Comparative analysis of radar point cloud density between the BSD and nuScenes, across different object categories. The histograms represent the number of radar points per annotated bounding box. For each class the density is evaluated for three different radial distance ranges measured in meters. The BSD demonstrates a significantly higher point density, offering finer detail for object detection. This comparison is made with all nuScenes radar point filters disabled to ensure a fair assessment, showcasing the superior radar resolution of BSD even when including potentially invalid points.}
    \label{fig:data_point_distribution}
\end{figure}
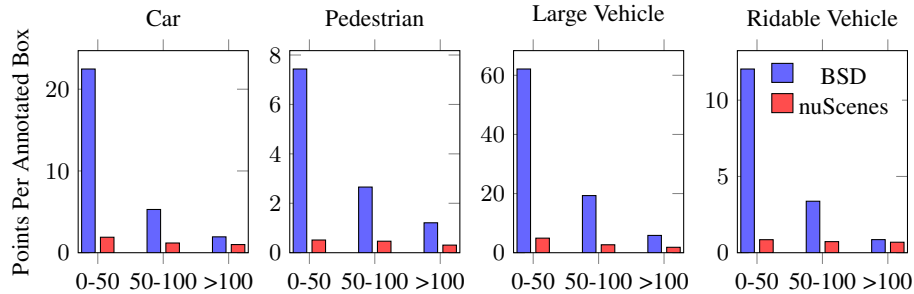

\subsection{Dataset Analysis}

The dataset was collected in various cities in Germany and the USA over the course of two years. 
For Germany, there are three main geospatial clusters: greater Stuttgart area (including cities like T\"{u}bingen, Reutlingen, Esslingen, Ulm, Heilbronn, and Karlsruhe),
Munich, and Berlin. For the US, we took measurements in California in the San Fransisco Bay Area.
While recording, interesting scenes were manually selected and labeled, as described in Sec.~\ref{sec:labeling}.

Diversity was the key design factor considered during the data collection and selection process. Here, diversity does not only refer to the geographical location of the collected scenes, but also to the road type (highway, rural, and urban), driving situation (straight road, intersection, roundabout, and tunnel), time of day (day time, twilight, and night) and weather conditions (clear, cloudy, and rainy). These different facets of diversity are labeled individually for each sequence as part of the provided metadata. 
Examples from the dataset are shown in Fig.\,\ref{fig:dataset_examples}. 
The camera is highly affected by lighting conditions, whereas radar and lidar maintain performance in the night scenarios. 

The dataset is divided into a train, validation, and test splits, whose data distribution is depicted in Fig.\,\ref{fig:data_piechart}. They were defined to enforce similar distributions with regards to the included classes, weather conditions, road types, and geographical locations. In addition, geographical integrity is maintained with no overlap occurring between the train/val/test splits. 

Fig.\,\ref{fig:data_range_vel_dist} depicts the distribution of the labels over range and velocity of the BSD in comparison to nuScenes. In BSD, labels are available up to high distances. Further, the distribution over the velocities indicates a significantly more diverse dataset, including also high velocities. This reflects the inclusion of different road situations such as highway and suburban scenes in contrast to the majority of prior datasets which focus only on urban situations as indicated in Tab.~\ref{tab:dataset_comparison}. 

A pivotal aspect of BSD is its inclusion of imaging radar sensors resulting in an enhanced radar point cloud density. Fig.\,\ref{fig:data_point_distribution} presents a stark comparison between BSD and nuScenes, elucidating the density of radar points within annotated bounding boxes for various object categories and distances. For that we projected the 3D point cloud to the 2D birds eye view (BEV) of the annotated boxes in the corresponding frame. Remarkably, BSD consistently exhibits a higher density of radar points across all categories. This increase in point density is critical as it translates into more detailed and accurate object representation, significantly benefiting radar-based detection algorithms, as will be empirically validated in the next section.

\section{Experiments}

\subsection{Baseline Model Performance}

\begin{figure}[t]
    \centering
    \input{figures/modality_baseline_models.tikz}
    \caption{Experimental results of single modality object detection models on nuScenes and on our dataset. For comparability, we evaluate all models up to a maximum distance of 50m from the ego vehicle to match the nuScenes leaderboard guidelines. We observe that the gap between radar and other modalities is significantly reduced in our work.}
    \label{fig:multimodal_results}
\end{figure}
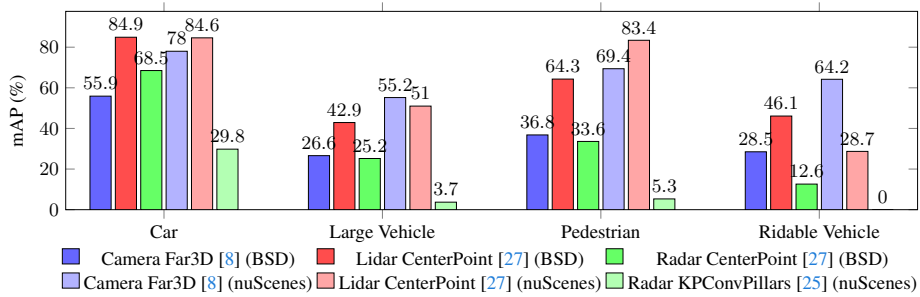

To establish benchmarks on BSD, we trained baseline models for object detection, employing Far3D~\cite{Jiang:2024} for camera data and CenterPoint~\cite{Yin:2021} for both radar and lidar data. For the radar model a 360\degree~point cloud is aggregated over 300ms, using multiple measurements from each sensor. These models were also trained on the nuScenes dataset to facilitate direct comparison, as illustrated in Fig.\, \ref{fig:multimodal_results}. Due to the extremely limited number of submissions within the public nuScenes leaderboards for radar, we compare against the KPConvPillars~\cite{Ulrich:2022} approach for this modality. Notably, BSD's imaging radar capabilities have markedly narrowed the performance gap between radar and other modalities, a disparity more pronounced in the nuScenes results. For instance, in BSD, the radar-based CenterPoint model approaches the performance of its lidar counterpart, particularly in detecting cars and large vehicles. This is reflected by the reduced mAP gap between the modalities of only 16.4\% and 17.7\% for the car and large vehicle classes, respectively, in comparison to 54.8\% and 47.3\% in nuScenes.


\subsection{Importance of Dataset Size}
\label{sec:experiment_dataset_size}

\begin{figure}[t]

    \begin{subfigure}[b]{0.99\textwidth}
        \input{figures/experiments_common_legend.tikz}
    \end{subfigure}
    
    \caption{Ablation studies of the radar object detection model. The left figure shows experimental results using subsets of the dataset. We observe that dataset size is important for detection performance. The right figure shows experimental results with different radar point cloud densities. We observe that high resolution radar sensors are crucial to detection performance.
    }
    \label{fig:experiments_ablation}
\end{figure}
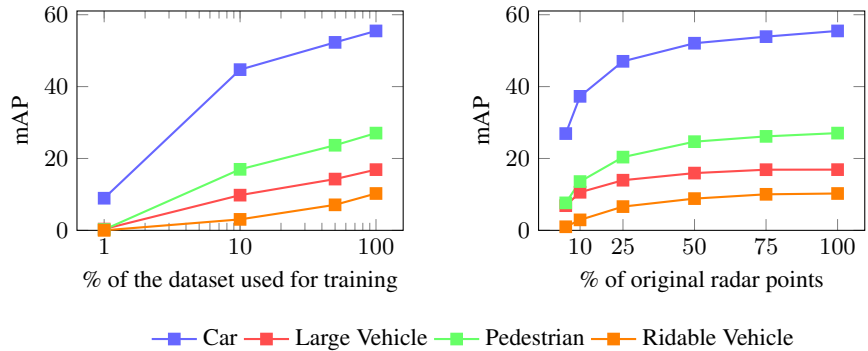

In Fig.\,\ref{fig:experiments_ablation}, we show the importance of dataset size by performing an ablation study on subsets of the full dataset. Specifically, we train the radar object detection CenterPoint model on subsets of the training sets and compute object detection metrics on the evaluation set. Those subsets are generated picking random samples until we cover the desired reduced percentage of the full dataset. The training then is performed on each of the subsets, while we still evaluate on the same test set as for the 100\% training. We observe that the performance decreases significantly as the size of the training set is reduced. Therefore, we show that the dataset size is essential.

\subsection{Importance of Radar Point Cloud Density}

In Fig.\,\ref{fig:experiments_ablation}, we show the advantages of higher resolution radars by sub-sampling the aggregated point clouds to fixed percentages of their original point count. We use the same architecture and training configuration as for the baseline radar object detection model but apply  a random sub-sampling algorithm to the training as well as the evaluation and test data. 
We observe that the detection performance decreases significantly over all classes with a decreasing point cloud density. We can therefore say that a high resolution is crucial in closing the gap between radar and lidar detection performance.



\begin{table}[t]
    \centering
    \resizebox{1.005\textwidth}{!}{%
        \begin{tabular}{@{}lccccccc@{}}
            \toprule
            \multirow{2}{*}{\shortstack{\vspace{0.12cm}\\Label type\\when training}} & 
            \multicolumn{4}{c}{mAP $\uparrow$} &&& \\
            \cmidrule{2-5}
            &
            LargeVehicle &
            PassengerCar &
            Pedestrian &
            RidableVehicle &
            mATE $\downarrow$ &
            mASE $\downarrow$ &
            mAOE $\downarrow$\\
            \midrule
            Manual & 10.00 (0.04) & 37.51 (0.04) & 14.20 (1.39) & 2.01 (0.12) & 53.07 (0.03) & 25.39 (0.01) & 71.61 (2.84) \\ \hdashline
            Auto & 12.12 (0.10) & 44.93 (0.34) & 18.58 (0.19) & 4.24 (0.05) & 47.84 (0.44) & 30.52 (0.06) & 73.63 (1.44) \\ \hdashline
            Manual + Auto & 13.77 (0.28) & 45.95 (0.42) & 19.85 (0.43) & 4.65 (0.07) & 46.50 (0.64) & 29.53 (0.02) & 70.99 (0.36) \\ \hdashline
            \shortstack{Auto + Manual\\(fine-tune)} & 13.26 (0.36) & 46.12 (0.44) & 19.76 (0.41) & 2.45 (0.04) & 46.32 (0.34) & 23.76 (0.1) & 56.19 (0.56) \\ \bottomrule
        \end{tabular}
    }
    \vspace{0.15cm} 
    \caption{Experimental results of a radar object detection model trained with varying labeling methods and evaluated with manual labels. Numbers in the table show the average of 5 separate runs, and variance values are enclosed in parentheses.}
    \label{tab:label_quality_results}
\end{table}

\subsection{Label Quality}



In this section we examine the performance of models that are trained on different quality levels of labels. We train radar-based models for the object detection task on manual, automatic, and both types of labels, while we test on manual labels only. The results are depicted in Tab.~\ref{tab:label_quality_results}. We observe a significant improvements by training on automatic labels compared to the smaller subset of the available manual ones. We explain this by the evident variability of the data that automatic label provides as well as the higher number of automatically labeled frames made possible by the lower requirements in terms of cost and time to generate them. A successful automatic labelling process enriches the training dataset significantly, enabling training on a large amount of data cost-effectively. Next, we train on manual and automatic labels simultaneously and observe more improvements in mAP and mean error metrics, demystifying the necessity of additional manual labels. Finally, we perform an additional experiment with both label methods where we fine-tune the model that was trained on automatic labels further on manual labels. We observe comparable mAP performance to the model trained on both simultaneously. However, the error metrics turned out significantly better.  \vspace{-4mm}


\section{Conclusion}

In conclusion, the Bosch street dataset (BSD) represents a significant advancement in the realm of autonomous driving datasets. By providing a large-scale, diverse, and multi-modal dataset that includes high-resolution imaging radar data alongside lidar and camera data, BSD addresses crucial gaps in currently available datasets. Our experimental results, which highlight reduced performance discrepancies between radar and other sensor modalities, underscore the potential of imaging radar in enhancing object detection and perception systems. The BSD not only sets a new precedent for sensor data quality and dataset scale but also serves as a foundational platform for the development of sophisticated HAD technologies. 

BSD is positioned as a strategic asset to foster academic and research collaborations with Bosch. This collaborative model aims to spark a surge in innovation and to accelerate progress in the autonomous driving domain. We are inviting for academic partnerships based on BSD and its associated assets that we believe will be instrumental in pushing the boundaries of what is possible in HAD research and development. Through these joint efforts, we are confident in our collective ability to advance towards the creation of more dependable and sophisticated autonomous driving systems, inching ever closer to the vision of fully autonomous vehicles.

%
%
\bibliographystyle{splncs04}
\bibliography{main-arxiv}

\end{document}

%% file: figures/dataset_statistics_piecharts.tikz
\begin{tabular}[t]{m{.3cm} m{2cm} m{2.3cm} m{2cm} m{2cm} m{2cm} m{2cm}}
    \rotatebox{90}{Train}&
    \begin{tikzpicture}
    \pie[color = {
            blue!60, 
            red!70},
        pos={0,0},
        hide number,
        radius=.6]
        {6.8/{},
        93.2/{}}
    \end{tikzpicture} &
    \begin{tikzpicture}
    \pie[color = {
            blue!60, 
            red!70,
            green!40,
            orange},
        pos={0,0},
        hide number,
        radius=.6]
        {
            75.0/{}, 
            6.7/{}, 
            12.8/{}, 
            5.4/{}
        }
    \end{tikzpicture} &
    \begin{tikzpicture}
    \pie[color = {
            blue!60, 
            red!70,
            green!40,
            purple},
        pos={0,0},
        hide number,
        radius=.6]
        {72.4/{},
        20.6/{},
        5.4/{},
        1.6/{}}
    \end{tikzpicture}
    &
    \begin{tikzpicture}
    \pie[color = {
            blue!60, 
            red!70,
            green!40,
            orange},
        pos={0,0},
        hide number,
        radius=.6]
        {48.7/{},
        25.0/{},
        23.6/{},
        2.4/{}}
    \end{tikzpicture}
    &
    \begin{tikzpicture}
    \pie[color = {
            blue!60, 
            red!70,
            green!40,
            orange},
        pos={0,0},
        hide number,
        radius=.6]
        {10.0/{},
        71.9/{},
        8.4/{},
        9.3/{}}
    \end{tikzpicture}
    &\\
    \rotatebox{90}{Validation}& 
    \begin{tikzpicture}
    \pie[color = {
            blue!60, 
            red!70},
        pos={0,0},
        hide number,
        radius=.6]
        {3.4/{},
        96.6/{}}
    \end{tikzpicture} &
    \begin{tikzpicture}
    \pie[color = {
            blue!60, 
            red!70,
            green!40,
            orange},
        pos={0,0},
        hide number,
        radius=.6]
        {
        75.8/{},
        6.8/{},
        12.8/{},
        4.6/{}
        }
    \end{tikzpicture} &
    \begin{tikzpicture}
    \pie[color = {
            blue!60, 
            red!70,
            green!40,
            purple},
        pos={0,0},
        hide number,
        radius=.6]
        {67.8/{},
        23.2/{},
        6.4/{},
        2.6/{}}
    \end{tikzpicture} &
    \begin{tikzpicture}
    \pie[color = {
            blue!60, 
            red!70,
            green!40,
            orange},
        pos={0,0},
        hide number,
        radius=.6]
        {48.8/{},
        23.9/{},
        24.3/{},
        2.6/{}}
    \end{tikzpicture}
    &
    \begin{tikzpicture}
    \pie[color = {
            blue!60, 
            red!70,
            green!40,
            orange},
        pos={0,0},
        hide number,
        radius=.6]
        {11.7/{},
        66.1/{},
        8.6/{},
        12.8/{}}
    \end{tikzpicture}
    &\\
    \rotatebox{90}{Test}& 
    \begin{tikzpicture}
    \pie[color = {
            blue!60, 
            red!70},
        pos={0,0},
        hide number,
        radius=.6]
        {16.0/{},
        84.0/{}}
    \end{tikzpicture} &
    \begin{tikzpicture}
    \pie[color = {
            blue!60, 
            red!70,
            green!40,
            orange},
        pos={0,0},
        hide number,
        radius=.6]
        {
        71.3/{},
        6.6/{},
        15.7/{},
        6.4/{}
        }
    \end{tikzpicture} &
    \begin{tikzpicture}
    \pie[color = {
            blue!60, 
            red!70,
            green!40,
            purple},
        pos={0,0},
        hide number,
        radius=.6]
        {69.6/{},
        20.4/{},
        7.6/{},
        2.4/{}}
    \end{tikzpicture} &
    \begin{tikzpicture}
    \pie[color = {
            blue!60, 
            red!70,
            green!40,
            orange},
        pos={0,0},
        hide number,
        radius=.6]
        {56.0/{},
        19.7/{},
        20.5/{},
        3.5/{}}
    \end{tikzpicture}
    &
    \begin{tikzpicture}
    \pie[color = {
            blue!60, 
            red!70,
            green!40,
            orange,},
        pos={0,0},
        hide number,
        radius=.6]
        {10.8/{},
        67.7/{},
        8.8/{},
        11.7/{}}
    \end{tikzpicture}
    &\\
    \multicolumn{2}{r}{
    \raisebox{-.2cm}{
    \begin{tikzpicture}
    \hspace{-1.25em}
    \pie[color = {
            blue!60, 
            red!70},
        pos={0,0},
        hide number,
        radius=0.0,
        text = legend]
        {39.6/{Manual label},
        60.4/{Auto-label}}
    \end{tikzpicture}}} &
    \hspace{-4em}
    \begin{tikzpicture}
    \pie[color = {
            blue!60, 
            red!70,
            green!40,
            orange},
        pos={0,0},
        hide number,
        radius=0.0,
        text = legend]
        {25/{Car},
        25/{Large Vehicle},
        25/{Pedestrian},
        25/{Ridable Vehicle}}
    \end{tikzpicture} &
    \hspace{-3em}
    \begin{tikzpicture}
    \pie[color = {
            blue!60, 
            red!70,
            green!40,
            purple},
        pos={0,0},
        hide number,
        radius=0.0,
        text = legend]
        {25/{Clear},
        25/{Cloudy},
        25/{Rainy},
        25/{Other}}
    \end{tikzpicture}&
    \hspace{-3em}
    \begin{tikzpicture}
    \pie[color = {
            blue!60, 
            red!70,
            green!40,
            orange},
        pos={0,0},
        hide number,
        radius=0,
        text = legend]
        {51.1/{Urban},
        33.8/{Rural},
        12.6/{Highway},
        2/{Tunnel}}
    \end{tikzpicture}
    &
    \hspace{-3em}
    \begin{tikzpicture}
    \pie[color = {
            blue!60, 
            red!70,
            green!40,
            orange},
        pos={0,0},
        hide number,
        radius=0,
        text = legend]
        {0.5/{Berlin},
        0.5/{BW},
        0.5/{SF},
        0.5/{Munich}}
    \end{tikzpicture}
    &\\
\end{tabular}

%% file: figures/labels_over_range.tikz
\begin{tikzpicture}
\begin{axis}[
    xlabel={range (m)},
    ylabel={label density},
    xlabel near ticks,
    ylabel near ticks,
    ymin=0,
    ymax=0.1,
    xmin=0,
    xmax=200,
    y tick label style={
        /pgf/number format/.cd,
        fixed,
        fixed zerofill,
        precision=2,
        /tikz/.cd
    },
    legend pos=north east,
    legend style={draw=none,},
    width=.99\textwidth,
    font=\footnotesize  
    ]
    
  \addplot[color=blue!60,thick,name path=bsd] table [x=BIN, y=DENSITY] {figures/labels_over_range_bsd.txt};

  \addplot[color=red!70,thick,name path=nusc] table [x=DISTANCE, y=LABELS] {figures/labels_over_range_ns.txt};

  \addplot[draw=none,name path=xaxis,domain=0:200] {0};
  \addplot[blue!60, opacity=0.05] fill between [of=xaxis and bsd, soft clip={domain=0:200}];
  \addplot[red!70, opacity=0.05] fill between [of=xaxis and nusc, soft clip={domain=5:200}];
\end{axis}
\end{tikzpicture}

%% file: figures/labels_over_velo.tikz
\begin{tikzpicture}
\begin{axis}[
    xlabel={velocity (m/s)},
    ylabel=label density,
    xlabel near ticks,
    ylabel near ticks,
    ymin=0,
    ymax=0.3,
    xmin=0,
    xmax=100,
    y tick label style={
        /pgf/number format/.cd,
        fixed,
        fixed zerofill,
        precision=2,
        /tikz/.cd
    },
    legend pos=north east,
    legend entries={BSD, nuScenes},
    legend style={draw=none,},
    width=.99\textwidth,
    font=\footnotesize  
    ]
    
  \addplot[color=blue!60,thick,name path=bsd] table [x=BIN, y=DENSITY] {figures/labels_over_velo_bsd.txt};

  \addplot[color=red!70,thick,name path=nusc] table [x=DISTANCE, y=LABELS] {figures/labels_over_velo_ns.txt};

  \addplot[draw=none,name path=xaxis,domain=4:80] {0};
  \addplot[blue!60, opacity=0.05] fill between [of=xaxis and bsd, soft clip={domain=4:80}];
  \addplot[red!70, opacity=0.05] fill between [of=xaxis and nusc, soft clip={domain=4:80}];
\end{axis}
\end{tikzpicture}

%% file: figures/statistic_point_cloud_density_test.tikz
\begin{tikzpicture}
    
    \begin{axis}[%
        name=plot1,
        width=0.315\textwidth,
        height=.35\columnwidth,
        ylabel={Points Per Annotated Box},
        title=Car,
        ybar,
        symbolic x coords={0-50,50-100,>100},
        ymin=0,
        enlarge x limits=0.15,
        bar width=0.18cm,
        ]
        \addplot[ybar,fill=blue!60] table [col sep=comma] 
        {statistics/statistic_point_cloud_density/bsd_car.txt};
        \addplot[ybar,fill=red!70] table [col sep=comma] 
        {statistics/statistic_point_cloud_density/nuscenes_car.txt};
    ]
    \end{axis}
    \begin{axis}[%
        name=plot2,
        at=(plot1.right of south east), anchor=left of south west,
        width=0.315\textwidth,
        height=.35\columnwidth,
        title=Pedestrian,
        ybar,
        symbolic x coords={0-50,50-100,>100},
        ymin=0,
        enlarge x limits=0.15,
        bar width=0.18cm,
        ]
        \addplot[ybar,fill=blue!60] table [col sep=comma] 
        {statistics/statistic_point_cloud_density/bsd_pedestrian.txt};
        \addplot[ybar,fill=red!70] table [col sep=comma] 
        {statistics/statistic_point_cloud_density/nuscenes_pedestrian.txt};
    \end{axis}
    \begin{axis}[%
        name=plot3,
        at=(plot2.right of south east), anchor=left of south west,
        width=0.315\textwidth,
        height=.35\columnwidth,
        title=Large Vehicle,
        ybar,
        symbolic x coords={0-50,50-100,>100},
        ymin=0,
        enlarge x limits=0.15,
        bar width=0.18cm,
        ]
        \addplot[ybar,fill=blue!60] table [col sep=comma] 
        {statistics/statistic_point_cloud_density/bsd_large_vehicle.txt};
        \addplot[ybar,fill=red!70] table [col sep=comma] 
        {statistics/statistic_point_cloud_density/nuscenes_large_vehicle.txt};
    ]
    \end{axis}
    \begin{axis}[%
        name=plot4,
        at=(plot3.right of south east), anchor=left of south west,
        width=0.315\textwidth,
        height=.35\columnwidth,
        title=Ridable Vehicle,
        ybar,
        symbolic x coords={0-50,50-100,>100},
        ymin=0,
        enlarge x limits=0.15,
        bar width=0.18cm,
        legend pos=north east,
        legend entries={BSD, nuScenes},
        legend style={draw=none,},
        ]
        \addplot[ybar,fill=blue!60] table [col sep=comma] 
        {statistics/statistic_point_cloud_density/bsd_bicycle.txt};
        \addplot[ybar,fill=red!70] table [col sep=comma] 
        {statistics/statistic_point_cloud_density/nuscenes_bicycle.txt};
    ]
    \end{axis}
    \begin{axis}[
        axis y line=none,
        axis x line=none,
        ticks=none,
        axis line style={draw=none}
        name=plot_invisible,
        at=(plot1.left of south west), anchor=left of south west,
        width=1.0\textwidth,
        height=.35\columnwidth,
        xlabel={Points Per Annotated Box},
        ]
    \end{axis}
\end{tikzpicture}

%% file: figures/modality_baseline_models.tikz
\begin{tikzpicture}[scale=.8]
\begin{axis}[
    ybar,
    width=1.29\columnwidth,
    height=.4\columnwidth,
    enlarge x limits=0.15,
    enlarge y limits={value=0.15,upper},
    legend style={draw=none, at={(0.5,-0.17)},
      anchor=north,legend columns=3},
    ylabel={mAP (\%)},
    ylabel near ticks,
    ymin=0,
    symbolic x coords={Car,Large Vehicle,Pedestrian,Ridable Vehicle},
    xtick=data,
    nodes near coords,
    nodes near coords align={vertical},
    bar width = 10pt,
    ]
    \addplot[ybar,fill=blue!60] coordinates{ 
        (Car, 55.9)
        (Large Vehicle, 26.6)
        (Pedestrian, 36.8)
        (Ridable Vehicle, 28.5)
    };
    \addplot[ybar,fill=red!70] coordinates{ 
        (Car, 84.9)
        (Large Vehicle, 42.9)
        (Pedestrian, 64.3)
        (Ridable Vehicle, 46.1)
    };
    
    \addplot[ybar,fill=green!60] coordinates{ 
        (Car, 68.5)
        (Large Vehicle, 25.2)
        (Pedestrian, 33.6)
        (Ridable Vehicle, 12.6)
    };
    \addplot[ybar,fill=blue!30] coordinates{ 
        (Car, 78)
        (Large Vehicle, 55.2)
        (Pedestrian, 69.4)
        (Ridable Vehicle, 64.2)
    };
    \addplot[ybar,fill=red!35] coordinates{ 
        (Car, 84.6)
        (Large Vehicle, 51.0)
        (Pedestrian, 83.4)
        (Ridable Vehicle, 28.7)
    };
    \addplot[ybar,fill=green!30] coordinates{ 
        (Car, 29.8)
        (Large Vehicle, 3.7)
        (Pedestrian, 5.3)
        (Ridable Vehicle, 0.0)
    };
    \legend{Camera Far3D \cite{Jiang:2024} (BSD), Lidar CenterPoint \cite{Yin:2021} (BSD), Radar CenterPoint \cite{Yin:2021} (BSD), 
    Camera Far3D \cite{Jiang:2024} (nuScenes), Lidar CenterPoint \cite{Yin:2021} (nuScenes), Radar KPConvPillars \cite{Ulrich:2022} (nuScenes)}
\end{axis}
\end{tikzpicture}

%% file: figures/experiments_common_legend.tikz
\begin{tikzpicture}
\begin{axis}[
    name=ax1,
    xlabel=\% of the dataset used for training,
    ylabel=mAP,
    xlabel near ticks,
    xmode=log,
    xtick={1,10,100},
    ylabel near ticks,
    ymin=0,
    width=.49\textwidth,
    height=4.5cm,
    legend style={draw=none, font=\small, at={(0.2,-0.4)},anchor=north west,legend columns=4},
    log ticks with fixed point,
    x tick label style={/pgf/number format/1000 sep=\,},
    ]
  \addlegendentry{Car}
  \addplot[color=blue!60,mark=square*,thick] coordinates {
    (1, 8.9393)
    (10, 44.7189)
    (50, 52.3038)
    (100, 55.4900)
  };
  \addlegendentry{Large Vehicle}
  \addplot[color=red!70,mark=square*,thick] coordinates {
    (1, 0.3650)
    (10, 9.7967)
    (50, 14.2675)
    (100, 16.8953)
  };
  \addlegendentry{Pedestrian}
  \addplot[color=green!60,mark=square*,thick] coordinates {
    (1, 0.2076)
    (10, 16.9812)
    (50, 23.6812)
    (100, 27.0606)
  };
  \addlegendentry{Ridable Vehicle}
  \addplot[color=orange,mark=square*,thick] coordinates {
    (1, 0.0000)
    (10, 3.0513)
    (50, 7.1017)
    (100, 10.2551)
  };
\end{axis}

\begin{axis}[
at={(ax1.south east)},
    xshift=1.8cm,
    xlabel=\% of original radar points,
    ylabel=mAP,
    xlabel near ticks,
    xtick={10, 25, 50, 75, 100},
    ylabel near ticks,
    ymin=0,
    width=.49\textwidth,
    height=4.5cm,
    legend style={font=\small, at={(1,-1)},anchor=south west,legend columns=4},
    x tick label style={/pgf/number format/1000 sep=\,},
    ]
  \addlegendentry{Car}
  \addplot[color=blue!60,mark=square*,thick] coordinates {
    (5, 26.9312)
    (10, 37.2892)
    (25, 47.0359)
    (50, 52.0673)
    (75, 53.9022)
    (100, 55.49)
  };
  \addlegendentry{Large Vehicle}
  \addplot[color=red!70,mark=square*,thick] coordinates {
    (5, 6.8669)
    (10, 10.5957)
    (25, 13.9515)
    (50, 15.9335)
    (75, 16.8778)
    (100, 16.8953)
  };
  \addlegendentry{Pedestrian}
  \addplot[color=green!60,mark=square*,thick] coordinates {
    (5, 7.6462)
    (10, 13.5755)
    (25, 20.3806)
    (50, 24.6760)
    (75, 26.1501)
    (100, 27.0606)
  };
  \addlegendentry{Ridable Vehicle}
  \addplot[color=orange,mark=square*,thick] coordinates {
    (5, 0.9975)
    (10, 2.9167)
    (25, 6.6087)
    (50, 8.8354)
    (75, 10.0284)
    (100, 10.2551)
  };
  \legend{}
\end{axis}
\end{tikzpicture}  

%% file: main-arxiv.bbl
\begin{thebibliography}{10}
\providecommand{\url}[1]{\texttt{#1}}
\providecommand{\urlprefix}{URL }
\providecommand{\doi}[1]{https://doi.org/#1}

\bibitem{Alibeigi:2023}
Alibeigi, M., Ljungbergh, W., Tonderski, A., Hess, G., Lilja, A., Lindstr{\"o}m, C., Motorniuk, D., Fu, J., Widahl, J., Petersson, C.: Zenseact open dataset: A large-scale and diverse multimodal dataset for autonomous driving. In: Proceedings of the IEEE/CVF International Conference on Computer Vision. pp. 20178--20188 (2023)

\bibitem{Bansal:2020}
Bansal, K., Rungta, K., Zhu, S., Bharadia, D.: Pointillism: Accurate 3d bounding box estimation with multi-radars. In: Proceedings of the 18th Conference on Embedded Networked Sensor Systems. pp. 340--353 (2020)

\bibitem{Barnes:2020}
Barnes, D., Gadd, M., Murcutt, P., Newman, P., Posner, I.: The oxford radar robotcar dataset: A radar extension to the oxford robotcar dataset. In: 2020 IEEE International Conference on Robotics and Automation (ICRA). pp. 6433--6438. IEEE (2020)

\bibitem{Bijelic:2020}
Bijelic, M., Gruber, T., Mannan, F., Kraus, F., Ritter, W., Dietmayer, K., Heide, F.: Seeing through fog without seeing fog: Deep multimodal sensor fusion in unseen adverse weather. In: Proceedings of the IEEE/CVF Conference on Computer Vision and Pattern Recognition. pp. 11682--11692 (2020)

\bibitem{Caesar:2020}
Caesar, H., Bankiti, V., Lang, A.H., Vora, S., Liong, V.E., Xu, Q., Krishnan, A., Pan, Y., Baldan, G., Beijbom, O.: nu{S}cenes: {A} multimodal dataset for autonomous driving. In: 2020 IEEE/CVF Conference on Computer Vision and Pattern Recognition (CVPR). pp. 11618--11628 (2020). \doi{10.1109/CVPR42600.2020.01164}

\bibitem{Deziel:2021}
D{\'e}ziel, J.L., Merriaux, P., Tremblay, F., Lessard, D., Plourde, D., Stanguennec, J., Goulet, P., Olivier, P.: Pixset: An opportunity for 3d computer vision to go beyond point clouds with a full-waveform lidar dataset. In: 2021 IEEE International Intelligent Transportation Systems Conference (ITSC). pp. 2987--2993. IEEE (2021)

\bibitem{Geiger:2013}
Geiger, A., Lenz, P., Stiller, C., Urtasun, R.: Vision meets robotics: The kitti dataset. International Journal of Robotics Research (IJRR)  (2013)

\bibitem{Jiang:2024}
Jiang, X., Li, S., Liu, Y., Wang, S., Jia, F., Wang, T., Han, L., Zhang, X.: Far3d: Expanding the horizon for surround-view 3d object detection. In: AAAI Conference on Artificial Intelligence (AAAI) (2024)

\bibitem{Kraus:2021}
Kraus, F., Scheiner, N., Ritter, W., Dietmayer, K.: The radar ghost dataset--an evaluation of ghost objects in automotive radar data. In: 2021 IEEE/RSJ International Conference on Intelligent Robots and Systems (IROS). pp. 8570--8577. IEEE (2021)

\bibitem{Lippke2023}
Lippke, M., Quach, M., Braun, S., K{\"o}hler, D., Ulrich, M., Bischoff, B., Tan, W.Y.: Exploiting sparsity in automotive radar object detection networks. arXiv preprint arXiv:2308.07748  (2023)

\bibitem{Ludloff:2002}
Ludloff, A.: {P}raxiswissen {R}adar und {R}adarsignalverarbeitung. Springer Vieweg (2002)

\bibitem{Mao:2021}
Mao, J., Niu, M., Jiang, C., Liang, H., Chen, J., Liang, X., Li, Y., Ye, C., Zhang, W., Li, Z., et~al.: One million scenes for autonomous driving: Once dataset. arXiv preprint arXiv:2106.11037  (2021)

\bibitem{Matuszka:2022}
Matuszka, T., Barton, I., Butykai, {\'A}., Hajas, P., Kiss, D., Kov{\'a}cs, D., Kuns{\'a}gi-M{\'a}t{\'e}, S., Lengyel, P., N{\'e}meth, G., Pet{\H{o}}, L., et~al.: {aiMotive} dataset: A multimodal dataset for robust autonomous driving with long-range perception. arXiv preprint arXiv:2211.09445  (2022)

\bibitem{Meyer:2019}
Meyer, M., Kuschk, G.: Automotive radar dataset for deep learning based 3d object detection. In: 2019 16th european radar conference (EuRAD). pp. 129--132. IEEE (2019)

\bibitem{Mostajabi:2020}
Mostajabi, M., Wang, C.M., Ranjan, D., Hsyu, G.: High-resolution radar dataset for semi-supervised learning of dynamic objects. In: Proceedings of the IEEE/CVF Conference on Computer Vision and Pattern Recognition Workshops. pp. 100--101 (2020)

\bibitem{Paek:2022}
Paek, D.H., Kong, S.H., Wijaya, K.T.: K-radar: 4d radar object detection for autonomous driving in various weather conditions. Advances in Neural Information Processing Systems  \textbf{35},  3819--3829 (2022)

\bibitem{Palffy:2022}
Palffy, A., Pool, E., Baratam, S., Kooij, J.F., Gavrila, D.M.: Multi-class road user detection with 3+ 1d radar in the view-of-delft dataset. IEEE Robotics and Automation Letters  \textbf{7}(2),  4961--4968 (2022)

\bibitem{Rebut:2022}
Rebut, J., Ouaknine, A., Malik, W., P{\'e}rez, P.: Raw high-definition radar for multi-task learning. In: Proceedings of the IEEE/CVF Conference on Computer Vision and Pattern Recognition. pp. 17021--17030 (2022)

\bibitem{Richards:2014}
Richards, M.A.: {F}undamentals of {R}adar {S}ignal {P}rocessing. McGraw-Hill, {S}econd edn. (2014)

\bibitem{ristic:2013}
Ristic, B., Vo, B., Vo, B., Farina, A.: A {T}utorial on {B}ernoulli {F}ilters: {T}heory, {I}mplementation and {A}pplications. IEEE Transactions on Signal Processing  \textbf{61}(13),  3406--3430 (July 2013). \doi{10.1109/TSP.2013.2257765}

\bibitem{Schumann:2021}
Schumann, O., Hahn, M., Scheiner, N., Weishaupt, F., Tilly, J.F., Dickmann, J., W{\"o}hler, C.: Radarscenes: A real-world radar point cloud data set for automotive applications. In: 2021 IEEE 24th International Conference on Information Fusion (FUSION). pp.~1--8. IEEE (2021)

\bibitem{Sheeny:2021}
Sheeny, M., De~Pellegrin, E., Mukherjee, S., Ahrabian, A., Wang, S., Wallace, A.: Radiate: A radar dataset for automotive perception in bad weather. In: 2021 IEEE International Conference on Robotics and Automation (ICRA). pp.~1--7. IEEE (2021)

\bibitem{Sun:2020}
Sun, P., Kretzschmar, H., Dotiwalla, X., Chouard, A., Patnaik, V., Tsui, P., Guo, J., Zhou, Y., Chai, Y., Caine, B., et~al.: Scalability in perception for autonomous driving: Waymo open dataset. In: Proceedings of the IEEE/CVF conference on computer vision and pattern recognition. pp. 2446--2454 (2020)

\bibitem{Svenningsson:2021}
Svenningsson, P., Fioranelli, F., Yarovoy, A.: Radar-pointgnn: Graph based object recognition for unstructured radar point-cloud data. In: 2021 IEEE Radar Conference (RadarConf21). pp.~1--6. IEEE (2021)

\bibitem{Ulrich:2022}
Ulrich, M., Braun, S., K{\"o}hler, D., Niederl{\"o}hner, D., Faion, F., Gl{\"a}ser, C., Blume, H.: Improved orientation estimation and detection with hybrid object detection networks for automotive radar. In: 2022 IEEE 25th International Conference on Intelligent Transportation Systems (ITSC). pp. 111--117. IEEE (2022)

\bibitem{Wilson:2023}
Wilson, B., Qi, W., Agarwal, T., Lambert, J., Singh, J., Khandelwal, S., Pan, B., Kumar, R., Hartnett, A., Pontes, J.K., et~al.: Argoverse 2: Next generation datasets for self-driving perception and forecasting. arXiv preprint arXiv:2301.00493  (2023)

\bibitem{Yin:2021}
Yin, T., Zhou, X., Kr{\"a}henb{\"u}hl, P.: Center-based 3d object detection and tracking. In: 2021 IEEE/CVF Conference on Computer Vision and Pattern Recognition (CVPR) (2021)

\bibitem{Zhang:2023}
Zhang, X., Wang, L., Chen, J., Fang, C., Yang, L., Song, Z., Yang, G., Wang, Y., Zhang, X., Li, J.: Dual radar: A multi-modal dataset with dual 4d radar for autononous driving. arXiv preprint arXiv:2310.07602  (2023)

\bibitem{Zheng:2022}
Zheng, L., Ma, Z., Zhu, X., Tan, B., Li, S., Long, K., Sun, W., Chen, S., Zhang, L., Wan, M., et~al.: {TJ4DRadSet}: A {4D} radar dataset for autonomous driving. In: 2022 IEEE 25th International Conference on Intelligent Transportation Systems (ITSC). pp. 493--498. IEEE (2022)

\bibitem{Zhou:2022}
Zhou, Y., Liu, L., Zhao, H., L{\'o}pez-Ben{\'\i}tez, M., Yu, L., Yue, Y.: Towards deep radar perception for autonomous driving: Datasets, methods, and challenges. Sensors  \textbf{22}(11), ~4208 (2022)

\end{thebibliography}
